# Structure and Optimization of Parameters for Neural Network Controllers in Automatic Control Systems

Sergey Feofilov
*Digital Control Systems Laboratory*
*Tula State University*
Tula, Russia
svfeofilov@mail.ru

Dmitry Khapkin
*Digital Control Systems Laboratory*
*Tula State University*
Tula, Russia
dima-hapkin@ya.ru

Andrey Kozyr
*Digital Control Systems Laboratory*
*Tula State University*
Tula, Russia
Kozyr_A_V@mail.ru

Eduard Heiss
*Digital Control Systems Laboratory*
*Tula State University*
Tula, Russia
edheiss73@gmail.com

Andrey Efromeev
*Digital Control Systems Laboratory*
*Tula State University*
Tula, Russia
age.sau@mail.ru

**Abstract**

The article outlines the methodology of structural and parametric synthesis of neural network controllers for controlling objects with limiters under incomplete information about the controlled object. Artificial neural networks are used to create controllers that are sequentially integrated into a control system with control objects. Reinforcement learning and pre-building a neural network imitator of the control object are used to synthesize the neural network controller. This approach is particularly effective when classical control system synthesis methods are not applicable due to significant nonlinearity and the difficulty in forming a mathematical model of the control object with the required accuracy. The proposed methods expand the class of technical systems for which direct synthesis of near-optimal control laws is possible. The robustness, adaptability and technical feasibility of neural network controllers make them interesting for practical applications. The main attention in the article is paid to the choice of neural network structure in the imitator and controller, formation of training samples taking into account the limitations of the control object.

## 1 Introduction

Currently, modern automatic control systems (ACS) for various technical devices are tending to increase the complexity of control objects and, consequently, their mathematical models. This is due to the use of new materials, technologies, and the need to expand previously unconsidered modes of operation. Additionally, most ACSs are implemented on digital computation systems because of their many advantages. However, this discreteness of signals in both time and level can lead to known problems, such as additional nonlinearities and delays that must be taken into account when synthesizing the system, and the possibility of auto-oscillations. Additionally, synthesized control systems should possess the properties of robustness and adaptability. As a rule, it is necessary to obtain control laws that are close to optimal according to a given criterion.

Classical synthesis methods are well-known, but their capabilities are significantly limited, especially for complex nonlinear control objects. For instance, consider an ideal automatic control system that is stable, adaptable to external conditions, and optimal according to some required criterion.

However, the object under consideration has many different nonlinearities, and its parameters are nonstationary. To achieve convenience, speed, and cost-effectiveness in development, it is recommended to implement this system on a digital computer with limited computing power. Developing such a control system requires the simultaneous use of methods from the theory of nonlinear digital systems, adaptive and robust systems, and synthesis of nonstationary and optimal control systems. Currently, constructing a workable synthesis technique based on this principle is not feasible. However, the relevance of the design of such systems requires the search for new approaches, algorithms and methods, which are able to cope with the indicated problem, at least in some cases. One promising approach is the utilization of methods from the field of artificial intelligence and machine learning, specifically, the use of artificial neural networks (ANN).

The synthesis and optimization of controllers for arbitrary nonlinear control systems is a broad issue. Therefore, this work focuses on control objects with various types of limiters. Such objects are prevalent in technical systems, including electric, hydraulic, and gas drives.

The choice of using an artificial neural network in the basis of controllers is due to the fact that when synthesizing real technical systems, the mathematical model is not accurate, there is a significant technological variation of parameters and it is not always possible to measure and use the necessary indicators of the system (phase variables) for control. All this leads to the deterioration of the quality of control laws, and thus the specific characteristics of the finished product. ANN are advantageous because they can be trained using signals obtained from real objects, which can then be used to form control within a system. These signals include input signals (control signals of the object) and output signals (signals from feedback sensors). ANNs can be adaptive due to real-time training.

## 2 Analysis of ANN application in automatic control systems

Artificial neural networks (ANNs) have traditionally been used in areas such as computer vision, machine learning, and virtual reality. Impressive achievements have been made in using ANNs for analyzing and synthesizing photo and video images, sound, and automatic translation systems. Recently, a variety of problems in robotics and mechatronics have been solved with the help of ANNs. The use of deep learning in robotics presents unique challenges and research questions that differ from those in other applications. Current research in this field focuses on applying neural networks and deep learning [1] to control complex dynamic systems, which can be categorized into several directions.

## 2.1 Synthesis of robust neural network controllers

An important direction in modern control theory is the synthesis of robust (insensitive to parametric deviations of the object) control. One of the most powerful approaches to the control of incompletely defined systems with constraints is a technique that uses the notion of an invariant control set, for which there are provable guarantees of the existence of an admissible control with feedback, capable of ensuring the finding of any closed trajectory of the system in a given set of admissible states [2]–[5]. Traditionally, this is achieved by relating the Lyapunov control function to an invariant set plan. A new approach to solving such problems is to use neurocontrollers. It is worth noting the work [6], which considers a methodology for the approximation-based synthesis of traditional stabilizing controllers for linear systems with polytopic uncertainty, including controllers with variable structure. The neural network architecture is based on the ReLU type activation function.

## 2.2 Identification of uncertainties in the process of operation

A common assumption in deep learning of a neural network controller is that the trained models are realized under closed set conditions [3],, i.e., the classes of signals encountered during operation are known and are exactly the same as those encountered during training. However, the control object often has to operate in constantly changing, uncontrolled real-world conditions, and they inevitably encounter instances of classes, scenarios, or environmental conditions that were not covered by the training data. The works [7], [8] are devoted to the study of uncertainty identification problems.

## 2.3 Selection of neural network architecture for neurocontroller

The main ANN architectures for application in neural network controllers are considered to be multilayer, recurrent and radial-basis neural networks. To date, typical schemes of ACS based on neural network controllers have been developed [9]. These schemes use in themselves only the architecture of multi-layer perceptron with input time delays [10]. Each of the typical schemes has its own drawbacks and is not universal. The designer of a neural network controller will have to conduct a study for the required control object to select a scheme for incorporating the controller into the control system and to choose the structure of the neural network to be used. As a result of the deep learning revolution, new activation functions have emerged, neural network training methods have been modified and refined, and new neural network architsectures have emerged. There is also active research in the field of reinforcement learning using deep ANNs [11]. In this method, the ANN learns to interact with the environment (virtual or real), receiving feedback from it about its actions (has feedback), in order to achieve a goal. This research is conducted mainly for learning the behavior of multi-agent systems.

## 2.4 Optimization tasks

There is also a large body of work on the use of neural networks to directly solve nonlinear programming problems [12]. The main idea of these methods is based on the use of ANN modeling the inverse dynamics of the object, which allows to reduce the nonlinear programming problem with constraints to a problem without constraints and makes it possible to significantly accelerate the optimization procedure. In [13], a sequential quadratic programming method is proposed using a Hopfield neural network model. The network is considered as a dynamic system converging to an optimal solution at time tending to infinity. There are known works, such as [14], in which a neural network is used to approximate the optimal controller (neural dynamic optimization). Offline, the neural network is tuned according to some quality indicator. Then the neural network needs to be retrained depending on the parametric changes of the control object. The trained neural network is then used in real time. There are also known works in which neural networks approximate table-defined nonlinear functions in order to obtain a differentiable function for the optimization procedure. In the works it is shown that the neural network allows to increase the smoothness of the obtained solutions in comparison with the use of splines.

Recent studies show that deep learning can be an effective tool to design optimal control systems for multidimensional nonlinear dynamical systems. But the behavior of these neural network controllers is still not fully understood. Designing optimal feedback controllers for multidimensional nonlinear systems remains a challenging and unsolved problem. Even assuming that the dynamics of the system is completely known, the design of such controllers requires solving a Hamilton-Jacobi-Bellman problem (PDE) whose dimensionality is the same as that of the state space. This leads to the well-known "curse of dimensionality" problem, precluding traditional discretization-based approaches. Some authors [15]–[17] use sparse grid interpolation to approximate the solution of the Hamilton-Jacobi-Bellman equation, called the value function, and its gradient, which is used to compute optimal feedback control. This direction has been further developed using nonlinear regression [11], [18] and sparse polynomials, which has significantly increased the maximum possible dimensionality of the problem. Alternatively, one can directly approximate the value gradient [17].

## 2.5 Stability of neural network control

Despite numerous examples of successful application of ANNs in the control of dynamic systems, neural network control systems still lack theoretical justifications (guarantees) of their stability, which may hinder their application in many safety-critical areas. The main explanation for this lies in their reputation as "black boxes" whose behavior cannot be predicted. Existing approaches to evaluating trained models are mainly based on testing using datasets. However, testing shows the presence, not the absence, of errors. If deep learning models are to be used, for example, in applications such as controlling autonomous aircraft, it is necessary to be able to test the safety-critical behavior of a closed-loop system.

A critical criterion that cannot currently be verified for neural network controllers is the stability assessment of the closed-loop system. Most current works attempt to use the direct Lyapunov method. There are various approaches to the numerical construction of Lyapunov functions for nonlinear systems with neural network control. Many of them are based on polynomial approximation of the dynamics and finding the sum of squares of polynomials as Lyapunov functions using semi-definite programming (SDP) [19]. This approach has been investigated in a large number of theoretical works, but in practice polynomial approximations impose significant limitations on the systems and the structure of the Lyapunov functions. Moreover, numerical sensitivity problems in SDP [20] are well known, which makes it very difficult to find solutions that fully satisfy the Lyapunov conditions.

Thus, research in the field of neural networks is very active and numerous theoretical and practical results have been obtained. Due to this, neural networks are widely used in many fields. However, their application as controllers in control systems is limited. The authors believe that we should fill this gap and use the potential of neural networks to control technical objects under conditions of incomplete information.

## 3 Structural synthesis of a closed-loop neural network control system

Next, we will consider forward propagation ANNs, for training which we will apply the method of backward error propagation through the forward neuroimitator [21]. This method uses two neural networks, one of which performs the function of the controller, and the second - the model of the control object, which is called neuroimitator. The training and control schemes are shown in Fig. 1.

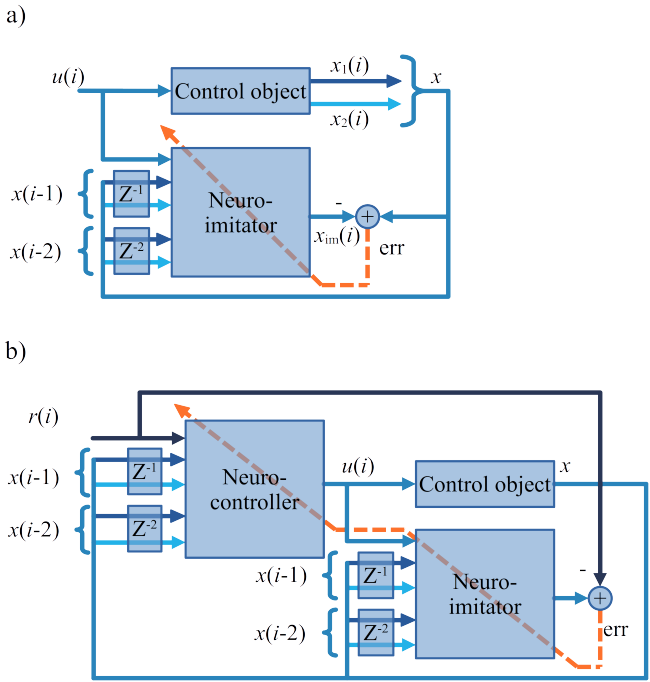

Fig. 1. Scheme for training the neurocontroller in the error back-passing method through the neuroimitator: a) training the neuroimitator, b) training the neurocontroller

Thus, the first task is to select the structure and training of the neuroimulator, which should simulate the behavior of the control object. In general, the structure of multilayer ANN is shown in Fig. 2.

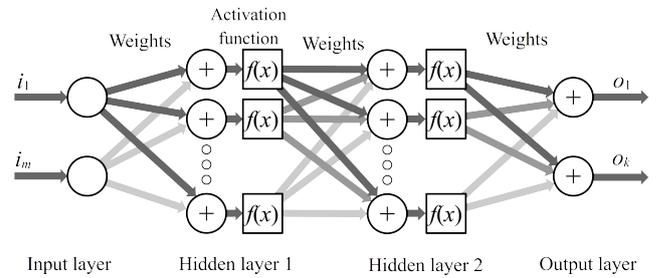

Fig. 2. Structure of a multilayer feed forward propagation neural network with two hidden layers

Today, the vast majority of ANNs are realized on digital devices, and for this reason we will consider them in discrete form. For a linear object, from the difference equation we can obtain the dependence for the direct dynamics ():

$$y(i) = \begin{aligned} & b_0 u(i) + b_1 u(i-1) + ... + \\ & + b_M u(i-M) - \\ & - a_1 y(i-1) - ... - a_K y(i-K), \end{aligned} \quad (1)$$

where $a_k$ and $b_m$ – are coefficients of the numerator and denominator of the Z-transfer function, respectively, $k = 0, 1 ... K$, $m = 0, 1 ... M$, $a_0 = 1$; $u$ is the signal at the input of the control object; $y$ is the output of the control object (one of the elements of the state vector $x$); $i$ is the current quantization cycle.

For the controller based on inverse dynamics, $u(i-1)$ is expressed from equation (), replacing $i$ by $(i+1)$ and $y(i+1)$ by $r(i)$:

$$u(i) = \frac{r(i) + a_1 y(i) + ... + a_K y(i-K+1)}{b_1} + \frac{-b_2 u(i-1) - ... - b_M u(i-M+1)}{b_1}. \quad (2)$$

where $r$ is the useful input signal.

Thus, for linear systems, the structure of the neuroimitator and neurocontroller can be chosen based on expressions () and (2), respectively, i.e., it is one neuron without activation function with weights corresponding to the coefficients of the difference equation.

In the following, the control objects with limiters such as saturation type limiters (Fig. 3) and rigid mechanical stop type limiters (Fig. 4) are considered.

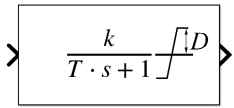

Fig. 3. Saturation-type limiter link

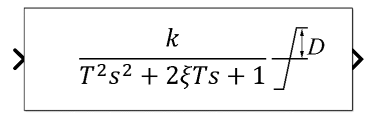

Fig. 4. Link with limiter type rigid mechanical stop

A link of the first type is described as follows:

$$\dot{x} = \begin{cases} \dfrac{(ku-x)}{T}, & \text{if } |x|<D \text{ or} \\ & |x|=D \text{ and } \dfrac{(ku-x)\,sign(x)}{T} \leq 0; \\ 0, & \text{if } |x|=D \text{ and } \dfrac{(ku-x)}{T} > 0. \end{cases} \quad (3)$$

The limit stop link of the rigid mechanical stop type has the following description:

$$\dot{x}_1 = x_2$$
$$\dot{x}_2 = \begin{cases} \dfrac{ku-x_1}{T^2} - \dfrac{2\xi x_2}{T}, & \text{if } |x_1|<D \text{ or} \\ & |x_1|=D \text{ and } sign(x_1)\leq 0; \\ 0, & \text{if } |x_1|=D \text{ and } (ku-x_1)sign(x_1)>0. \end{cases} \quad (4)$$

It is assumed that the impact with the stop is absolutely inelastic and the descent from the stop is continuous:

$$\begin{aligned} x_1(t^*+0) &= x_1(t^*-0), \\ x_2(t^*+0) &= 0, \end{aligned} \quad (5)$$

where $t^*$ is the time at the moment of impact.

To simulate and control such objects, it is proposed to use neural networks with the ReLU activation function [22], which is described by the following equation:

$$f(x) = \begin{cases} x, & x \geq 0; \\ 0, & x < 0; \end{cases} \quad (6)$$

Based on equations (3)–(6), the authors have developed neural network structures that repeat the discrete dynamics of objects with limiters. For example, for a link with a limiter of the rigid mechanical stop type, such a structure is shown in Fig. 5.

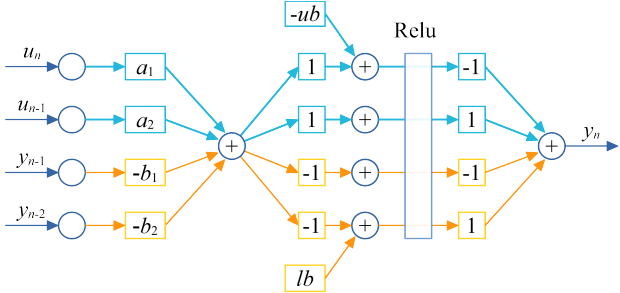

Fig. 5. Structure of a neuroimitator for a link with a limiter type rigid mechanical stop

In Fig. 5: $a_1$, $a_2$, $b_1$, $b_2$ are the values of the coefficients of the difference equation (1), $ub$ and $lb$ are the value of the upper and lower constraint, respectively.

Generalization leads to the following dependence for the minimum size of the hidden layer of the neural imitator:

$$n_{hiden} = 4 n_{sat} + 2 n_{lin}, \quad (7)$$

where $n$ is the order of the system, $n = n_{sat} + n_{lin}$, $n_{sat}$ is the number of constrained phase variables of the state vector, $n_{lin}$ is the number of unconstrained phase variables of the state vector.

Depending on the possibility of measuring phase variables, the structure of the neural network can be formed either on the basis of feedbacks constituting the full state vector of the control object, or on the basis of delays introduced in the feedback on the controlled variable. In the first case, the minimum network structure is as follows:

- 1st layer (input) dimension $n+1$ (object state vector and control signal);
- 2nd layer (hidden) – $4 n_{sat} + 2 n_{lin}$;
- 3rd layer (output) is of dimension n (state vector at the new step).

When using output signal delays:

- 1st layer (input) – $2n$, 2n consisting of the control signal of the object (1) and its delayed values ($n$-1), feedback from the control object (1) and its delayed values ($n$-1);
- 2nd layer (hidden) – $4 n_{sat} + 2 n_{lin}$;
- 3rd layer (output) – 1 (scalar output of the system).

Figs. 6, 7 show the results of comparing the dynamics of the link with a rigid mechanical stop type limiter with a neuroimitator when operating in the linear zone and reaching the limiters, respectively. The structure of the neuroimitator was chosen according to the developed algorithm, i.e., 3 neurons at the input and 2 neurons at the output. The hidden layer has 6 neurons with ReLU activation function.

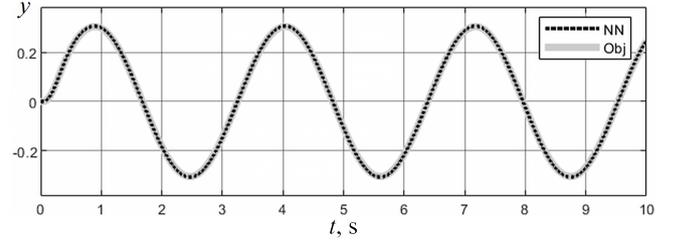

Fig. 6. Object and neuroimitator output without reaching the limiters.

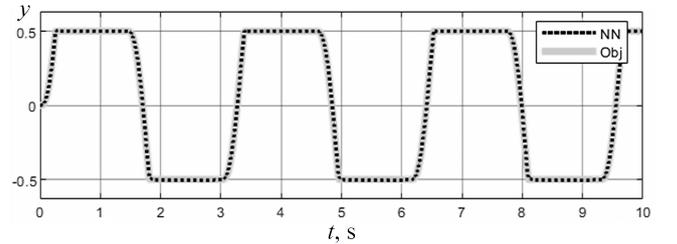

Fig. 7. Object and neuroimitator output with reaching the limiters

In Figs. 6 and 7 the dashed line indicates the output of the neuroimulator, the solid gray line shows the output of the control object. The figures show that the imitator with the proposed structure repeats the dynamics of the object with near-zero error (of the order of $10^{-15}$).

It is impossible to find a universal description of the neurocontroller structure for an arbitrary object. In general, a forward propagation ANN requires $2n$ inputs, where $n$ is the order of the control object. In addition, the following considerations are proposed: the order of the object determines the minimum size of the input and the number of delays for feedbacks, and for hidden layers the rule of step-by-step increase of neurons and the number of layers should be used until the result corresponding to the given requirements is achieved.

## 4 Formation of training sample for neurocontrollers in closed-loop systems with limiters

The method of error back propagation will be used to train neurocontrollers [23]. Based on the accumulated experience, the following recommendations are proposed to form a set of input signals (training sample) for the synthesis of neuroimitators and neurocontrollers for objects with limiters.

1. Signals should reflect the peculiarities of the object operation both in transient and steady-state modes. In general, typical continuous and discontinuous signals (sine and meander of different amplitudes and frequencies) should be used. The meander allows you to obtain information about the object's reaction to a jump change and constant value of the input. Harmonics are used to collect data on the processing of continuous, dynamically changing signals. Training sequences should be divided into two categories: limiters in the control object are not reached; limiters are reached, and if there are several limiters in the object, they are reached each separately and together. In this case, it is desirable that the number of data in the categories in the training sample be approximately equal.

2. It is necessary to record the response of the control object to all selected signals.

3. The collected data should be prepared. For forward propagation networks, the recorded signals are divided into pairs so that they represent the dependence of input data and result at one quantization clock, i.e. for the neuroimulator ($nn_{imit}$): $x_i = nn_{imit}(u_i, x_{i-1}, x_{i-2}, ...)$, and for the neurocontroller ($nn_c$): $u_i = nn_c(r_i, x_{i-1}, x_{i-2}, ...)$, where $i$ is the number of the discrete step, u is the object control signal, x is the object state, $r$ is the useful input signal (see Fig. 1). Thus, the collected pairs do not depend in any way on each other and on the signal from which they were derived, which allows to speed up training due to parallel computations.

## 5 Synthesis of a neurocontroller based on back propagation of error through a neuroimitator

Based on the above, the following algorithm for synthesizing a neurocontroller for objects with limiters is proposed.

1. The structure of the neurocontroller is selected, i.e., the types of layers, the number of neurons in each layer, and the activation functions are determined according to the formulated recommendations.

2. A training sample of data reflecting the dynamics of the control object operation is formed in accordance with the above considerations. The resulting training sample should be divided into two parts: a training sample and a test sample. Usually the sample is divided in the proportion of 90% to 10% respectively [24].

3. The neuroimitator is trained using the error back propagation method. A universal method is to divide the data into mini-batches using variations of adaptive stochastic gradient descent, for example, the Adam algorithm [25].

The mean square error is usually used as a loss function (MSE-loss) (8)

$$loss(a, \tilde{a}) = \frac{1}{k} \sum_{i=1}^{k} (a_i - \tilde{a}_i)^2, \quad (8)$$

where $a$ and $\tilde{a}_i$ are vectors with input (obtained as a result of ANN calculation on data from the training sample) and target values, respectively, $k$ is the number of values in the training sample.

The training epochs should be repeated until function (8) reaches the target value, which is determined by the neurocontroller designer.

4. The structure of the neurocontroller is determined and its training by the method of error back propagation is performed. At each quantization cycle the control value at the output of the neural network controller is calculated on the basis of the setting signal for the system and feedbacks from the object. The result of calculations is the value of the control signal, based on which the neuroimitator determines the new state of the EI. Then the obtained mismatch (8) is transmitted through the neuroimitator and neurocontroller by the method of error back propagation. In this case, only the neurocontroller is trained.

5. The quality of the obtained system in different modes of operation is evaluated. Since the training takes place in real time, it is possible to perform additional training of both the neuroimitator and the neurocontroller in case of deterioration of the control quality (e.g., when the parameters of the control object change). Thus, the system obtains the adaptability property, which, however, can be used only for slowly changing parameters. This is due to the fact that the training process requires large computational resources.

### 5.1 Example

Based on the developed algorithms, a neural network controller for the tracking hydraulic drive was synthesized, the simplified functional model of which is presented in Fig. 8.

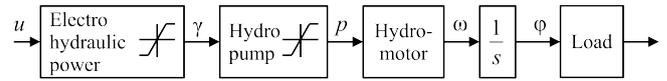

Fig. 8. Structure-functional diagram of the hydraulic drive

In Fig. 8: γ – angle of rotation of the hydraulic pump cradle; $p$ – pressure drop in the hydraulic motor; ω – speed of rotation of the output shaft; φ – angle of rotation of the shaft. The structural scheme takes into account inherent in the hydraulic drive limiter on the angle of rotation of the pump cradle: $|\gamma| \leq D_1$ and inertia-free limiter of the oil pressure drop in the power hydraulic lines: $|p| \leq D_2$.

The exact mathematical model for building a neurocontroller is generally not required, but an assumption must be made about its order.

Based on the developed algorithms, the following neural network structures were chosen (Fig. 9). If we assume that the EI is described by a system of nonlinear differential equations of the fourth order, the neuroimitator minimally requires five neurons at the input, four neurons without activation functions in the first hidden layer, 12 neurons with ReLU activation function and bias weights in the second layer, and four neurons in the output layer. For the neurocontroller, a forward propagation network with four inputs (state vector), two hidden layers was chosen: 24 neurons in the first layer, 8 neurons with leakyReLU activation functions in the second layer, and one neuron forming a signal to the control object at the output.

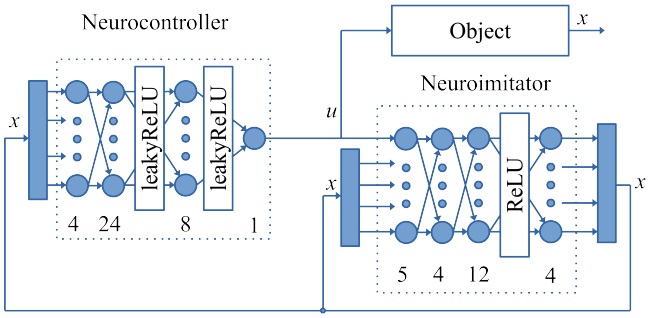

Fig. 9. Structure of the neural network system used in training the neurocontroller

For comparison with the synthesized neural network controller, an LQR [26] was designed for the hydraulic drive model without considering the limiters. The simplified hydraulic drive model presented in Fig. 10 was considered as a model.

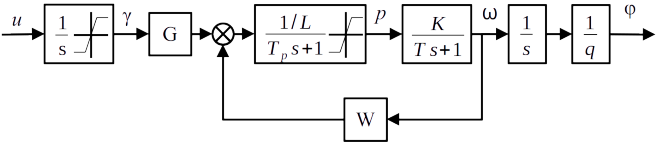

Fig. 10. Basic hydraulic drive model

In Fig. 10: $G$ is the flow rate gain of the variable displacement pump, $L$ is the total fluid leakage coefficient; $T_p = L * E/V$, where $E$ is the reduced value of the bulk modulus of the fluid, $V$ is the volume of fluid in the high pressure line.

In Fig. 10, the autonomous systems with the synthesized neural network controller and with LQR from a non-zero initial state (1 rad and 50 rad/c) are compared.

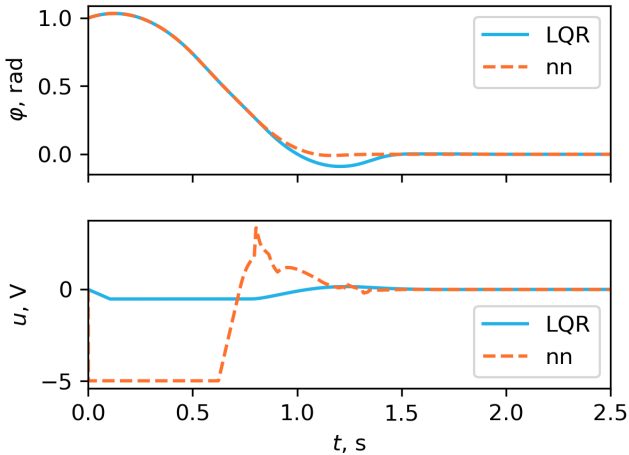

Fig. 11. Transition process from a non-zero initial state

From the simulation results, it can be seen that the system with synthesized neural network controller has a shorter transient time (1.86 s for the system with LQR and 1.32 s for the system with neural controller). At the same time, the overshoot is 10 times less. Thus, a controller capable of forming a nonlinear control law that allows stabilizing the system for a near-optimal control time and a Lyapunov function that guarantees asymptotic stability in the local region of phase space has been synthesized.

## 6 Conclusion

The analysis of the results indicates that in certain cases, the utilization of artificial neural networks (ANN) in closed-loop tracking automatic control systems can simplify and formalize the task of controller synthesis. The existing types of ANN and known methods of synthesis of neural network control systems are analyzed. Direct propagation neural networks were selected as the neural network controllers for control objects with limiters. The backpropagation method was used to train these networks. The use of ReLU or leakyReLU as activation functions is recommended.

An algorithm of structure formation of neurocontrollers and neuroimitators for the control object with limiters is developed, which allows determining the minimum number of layers and neurons in them. The input and output layer sizes are clearly defined during the synthesis of the neurocontroller. To achieve the desired result, the rule of step-by-step increase of neurons and layers should be used for the hidden layers. A method for forming a training sample algorithm has been developed, taking into account limiters in the control object. Additionally, a proposed method for synthesizing a neurocontroller allows for the determination of the main parameters of neural networks and training algorithms. The results obtained from this research can be used to form a method of synthesis and optimization of neural network control systems for controlling dynamic objects with limiters under conditions of incomplete information. The application of this methodology is demonstrated using a specific technical object as an example.

### Acknowledgment

This work was financially supported by the Ministry of Science and Higher Education of the Russian Federation under the state assignment FEWG-2022-0003.